# Predicting Defects in Laser Powder Bed Fusion using *in-situ* Thermal Imaging Data and Machine Learning

Sina Malakpour Estalaki[1], Cody S. Lough[2], Robert G. Landers[1], Edward C. Kinzel[1,3], Tengfei Luo[1,4, *]

[1]Department of Aerospace and Mechanical Engineering, University of Notre Dame, Notre Dame, Indiana, 46556, USA

[2]Department of Mechanical and Aerospace Engineering, Missouri University of Science and Technology, Rolla, MO, 65409, USA

[3]Department of Electrical Engineering, University of Notre Dame, Notre Dame, Indiana, 46556, USA

[4]Department of Chemical and Biomolecular Engineering, University of Notre Dame, Notre Dame, Indiana, 46556, USA

[*]Corresponding author: tluo@nd.edu

## Abstract

Variation in the local thermal history during the laser powder bed fusion (LPBF) process in additive manufacturing (AM) can cause microporosity defects, which add to the uncertainty of the properties of the built materials. *in-situ* sensing has been proposed to monitor the AM process to minimize defects, but the success requires establishing a quantitative relationship between the sensing data and the porosity, which is especially challenging for a large number of variables. Physics-based modeling can simulate such a relationship, but they are computationally costly. In this work, we develop machine learning (ML) models that can use *in-situ* thermographic data to predict the microporosity of LPBF stainless steel materials. This work considers two identified key features from the thermal histories: the time above the apparent melting threshold ($\tau$) and the maximum radiance ($T_{max}$). These features are computed, stored for each voxel in the built material, are used as inputs. The binary state of each voxel, either defective or normal, is the output. Different ML models are trained and tested for the binary classification task. In addition to using the thermal features of each voxel to predict its own state, the thermal features of neighboring voxels are also included as inputs. This is shown to improve the prediction accuracy, which is consistent with thermal transport physics around each voxel contributing to its final state. Among the models trained, the F1 scores on test sets reach above 0.96 for random forests. Feature importance analysis based on the ML models shows that $T_{max}$ is more important to the voxel state than $\tau$. The analysis also finds that the thermal history of the voxels above the present voxel is more influential than those beneath it. Our study significantly extends the capability to use *in-situ* thermographic data to predict porosity in LPBF materials. Since ML models are fast, they may play integral roles in the optimization and control of such AM technologies.

# 1. Introduction

Additive Manufacturing (AM) is recognized as a new paradigm for the manufacturing industry. It stands out due to its capability of creating complex, multi-material, and multi-functional designs and unique position in advancing manufacturing through data and machine intelligence [1]. Recently, data-driven machine learning (ML) techniques are being applied to various AM applications to monitor building processes, detect defects or anomalies, and enhance decision-making leveraging data collected through different sensors [2]. Jin *et al.* reviewed different ML methods that are used to systematically optimize different stages of AM processes, ranging from geometrical design and process parameter configuration to *in-situ* anomaly detection [1]. Wang *et al.* also provided a comprehensive review on the state-of-the-art of ML applications in a variety of AM domains [3].

Detection of the defects due to variation in process conditions is of prime importance for quality control of AM. Kadam *et al.* used different ML algorithms in combination with pre-trained convolutional neural networks for fault detection in a fused deposition modelling-based 3D printing process in a layer-by-layer manner, where layers were treated as images [4]. Li *at al.* proposed a ML scheme to detect bumps on the surfaces of AM samples by describing them using point clouds and detecting anomalies in the distances among the points [5]. Jin *et al.* used layer-by-layer optical images and convolutional neural networks to distinguish imperfections for transparent hydrogel-based bio-printed materials [6]. These prior studies focused on obvious defects that are visible in optical images where the optical images are also inputs for the ML models. However, in some other AM techniques, like laser powder bed fusion (LPBF), defects can be in the microscale which *in-situ* optical sensors cannot detect easily. In addition, using images for defect detection does not link the defect origin to process conditions and thus cannot be used for informing process optimization or feedback control.

Micropores in LPBF materials are usually detected using high-resolution micro-computer tomography (μ-CT) [7,8], but each scan can take hours, making it infeasible for *in-situ* defect detection. More advanced techniques like the synchrotron radiation micro-tomography [9,10] are capable of high-resolution imaging of pores in LPBF materials, but they need highly specialized facility that are not easily applicable to wide-spread development. High-speed x-ray [11,12] can image pore dynamics *in-situ* during the LPBF process, but it also needs specialized facilities and can only monitor a relatively small volume of a few hundreds of micrometers in size.

However, the origin of the micropores is rooted in the thermal history the material has experienced during the LPBF process [9,13]. Using thermal features that can be monitored *in-situ* to predict microporosity is thus attractive [14], and there have been several studies exploring their relationships. Scime and Beuth monitored the melt-pool using high-speed camera in the LPBF of stainless steel and used unsupervised learning and computer vision techniques to distinguish normal and abnormal melt-pools from the optical images [15]. It is desirable to further establish quantitative models to link the melt-pool information to porosity. Cobert *et al.* established such a relation using the support vector machine ML model to classify if a printed region is defective or nominal by training the model against *in-situ* optical images with *ex-situ* CT data. They achieved a defect detection accuracy, defined as (true positive + true negative)/total population, greater than 80% as demonstrated using cross-validation [16]. However, since the number of defective regions is usually much smaller than that of normal regions in common LPBF materials, this accuracy of ~80% is not particularly high. This was indicated by the true positive and true negative rates, which were both ~60% at best. That is to say, their model had high probability to predict false positive and false negative labels.

Using optical cameras to monitor melt-pool is convenient, but it offers in-direct information for the thermal history of the LPBF process. Directly sensing the thermal features may provide more direct information for defect prediction. Baumgartl *et al.* used *in-situ* thermographic imaging to detect printing defects by training convolutional neural network (CNN) ML models, achieving an overall accuracy of 97.9% and true positive and true negative rates both above 96.8%. However, the detected defects were delamination and splatter, which are much larger defects than micropores and easier to detect in nature [17]. Paulson *et al.* monitored the surface temperature histories of a Ti64 LPBF process *in-situ* by an IR camara and used statistical ML models to correlate the thermal history to the keyhole porosity formation in the subsurface [13]. They extracted different features from the temperature histories and formulated the problem into a binary classification task to determine if subsurface pores would form or not. In this study, the pores were identified in the melt pool as circular regions with lighter intensity than other solid regions using high-speed x-ray *in-situ*. From the x-ray, they were able to distinguish small pores from large pores, which can have different origins, but they found that ML models perform better when not distinguishing between small and large pores. Trained by 112 temperature histories, the cross-validation accuracy using the leave-one-out scheme was shown to be 83.9-89.3% depending on the different ML models and features used. On the test set, which contained 8 temperature histories and the corresponding porosity information, the ML model accuracy was 87.5-100.0% for different models and features used. However, *in-situ* high-speed x-ray is a very specialized tools not widely accessible, and the overall data size was relatively small for ML. Around the same time, Lough *et al.* used short-wave infrared (SWIR) imaging to monitor the spatial and temporal thermal features of the whole LPBF process of stainless steel, voxel by voxel [18]. They showed that various thermal features extracted from the SWIR data for each voxel could be indicative of whether the voxel contains pore defects, which were characterized by *ex-situ* μ-CT. In a follow-up study, Lough *et al.* found that two thermal features are most influential for defect detection, including the time above the apparent melting temperature and maximum radiance. In this study, they used linear regression to correlate these two thermal features to the porosity in a voxel-by-voxel manner and obtained area under curve (AUC) scores up to ~0.86 of the receiver operating characteristic (ROC) curve [19]. However, as shown by Paulson *et al.* [13], pores can form in the subsurface of the present top layer, and thus including thermal features surrounding a voxel, instead of only those of itself, may be needed to accurately predict the existence of pores.

In this work, we reduce thermal time histories recoded by an *in-situ* short-wave camera to take the two proven important thermographic features extracted from our previous work [18], including time above apparent melting threshold ($\tau$) and the maximum radiance ($T_{max}$), and develop ML models to predict the microporosity of LPBF stainless steel materials in a voxel-by-voxel manner. The microporosity is characterized using *ex-situ* μ-CT and voxels are labeled as either defective or healthy, depending on the density of micropores in each voxel. Different ML models are trained and tested for this binary classification task. Besides using the thermal features of each voxel to predicts its own state, those of neighboring voxels are also included, which is shown to improve the prediction accuracy. Such a finding suggests that the thermal history around a voxel can also impact its state due to heat transfer. Among the models trained, the F1 score on the hold-out test sets reaches above 0.96 for random forests (RF) models. Feature importance analysis based on the ML models shows that $T_{max}$ is more important to the voxel state than $\tau$. The analysis also finds that the thermal features of the voxels above the present voxel is more influential than those beneath it, correctly reflecting the influence of thermal history. Our study demonstrates the viability to use *in-situ* thermographic data to predict porosity in LPBF materials in high precision. Since ML models are fast, they may play integral roles in the optimization and control of such AM technologies.

## 2. Data and Models

Introduce experiment. A 304L stainless steel cylindrical part with a diameter of 4 mm and a height of 20 mm was fabricated on a Renishaw AM 250 selective laser melting machine. Other manufacturing details, such as layer height, laser path, and sensing equipment can be found in Ref. [18]. The two thermal features (*i.e.*, $\tau$ and $T_{max}$) for each voxel of the built material are used as inputs, while the binary labels characterizing if the voxel is defective (*i.e.*, porous) or normal (*i.e.*, fully dense) are outputs for the ML model. The voxel-wise binary labels are obtained by registering the part's thermal feature voxel reconstruction with its post-processed μ-CT data. In the post-processing, the μ-CT data is first down-sampled from 15×15×10 μm$^3$ per voxel to the thermal features' 130×135×50 μm$^3$ voxel size. The down-sampling is performed by defining a grid with the larger voxel size and then using the grayscale intensity values of the smaller voxels encompassed by a larger voxel to determine both a new grayscale voxel value by averaging and a binary state by the percentage of porosity present. This produces both a grayscale intensity μ-CT-based reconstruction and binary μ-CT-based reconstruction for the part, where the binary states are defective and normal. A binary voxel is considered as porosity if more than 5% of the original smaller voxels within the new voxel volume correspond to porosity. This small percentage allows the lower resolution binary map to capture the smaller porosity features contained in the full resolution μ-CT data. Next, the down-sampled grayscale μ-CT reconstruction is registered with the thermal feature reconstruction. The data sets are registered in the *z*-direction (*i.e.*, build direction) by a manual translation and then registered automatically in the *x-y* plane by translations determined through a multimodal intensity-based algorithm. The translations from the registration step are then applied to the binary µ-CT-based reconstructions. This finally provides the binary labels for the thermal feature voxels because the porosity state is now known from the corresponding point in the binary reconstruction for a given (*x*, *y*, *z*) coordinate.

There are a total of 398 layers with each layer consisting of around 465 voxels arranged in a cylindrical shape (**Figure 1a**). For each voxel, we used the thermal features of itself and those of its neighbors as inputs to predict its binary state. **Figure 1b** shows an example schematic of including the 1st nearest neighbors for the voxel state prediction and how the neighboring voxels are indexed. In this example, the total number of voxels involved in predicting the state of the central voxel is 3×3×3, and since the neighbor information is processed using sliding window kernels, we call this case K3 (i.e., kernel 3×3×3). **Figure 1c** shows a two-dimensional illustration of the windowing kernel. We have also tested 1×1×1 (no neighbors), 5×5×5 (up to 2nd nearest neighbors) and 7×7×7 (up to 3rd nearest neighbors) kernel cases in this study, and they are respectively called K1, K5 and K7 cases. Voxels near the top, bottom and side surfaces of the built volume are excluded from the data if they do not have the complete set of neighbors (see right panel in **Figure 1c**). The thermal features are then put into a one-dimensional vector with the first half the elements including $\tau$ of voxels and the second half containing $T_{max}$ (**Figure 1d**). The inputs are scaled according to $x' = \frac{x - x_{min}}{x_{max} - x_{min}}$, where $x_{min}$ and $x_{max}$ are the min and max of each feature. It is noted that the current data is a highly imbalanced with 92% of all the voxels being normal (labeled as 0) and only 8% defective (labeled as 1). To improve the ML model accuracy, we employ the Borderline-Synthetic Minority Oversampling (Borderline-SMOTE) technique [20] to duplicate samples in the minority class so that the positive and negative labels are balanced to 1:1 for the training process. Although oversampling is used for training, we note that all the model accuracies reported are quantified on the unbalanced original data.

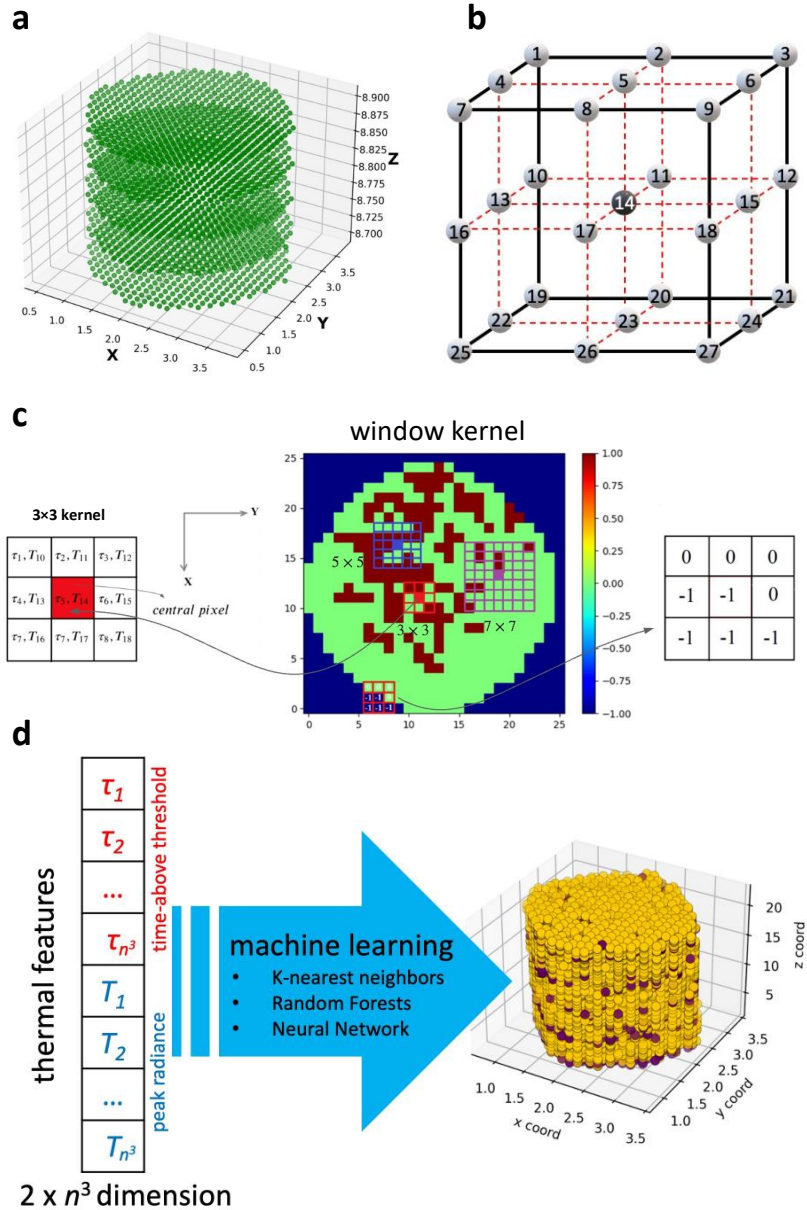

**Figure 1.** (a) Illustration of the voxelated built volume (for clarification, only five layers are illustrated). (b) An example of the neighboring voxels (white) for a central voxel (black), and the indexing convention used. (c) Illustration of using sliding window kernels with different dimensions to extract the neighbors for each voxel, and how the surface

voxels are detected (right panel), which are excluded in the ML tasks. **(d)** The thermal features of neighboring voxels are put into a one-dimensional vector as inputs for ML models to predict the state (defective – 1; normal – 0) of each voxel.

Six different ML classification models, including K-nearest neighbors (KNN), Random Forests (RF), Decision Trees (DT), Multi-Layer Perceptron (MLP), Logistic Regression (LR) and AdaBoost, are trained and tested. Certain percentages (7%, 10%, 20%, 30% and 40%) of data are held out for testing to evaluate the predictivity of the models, while the rest are used for training. The hyperparameters used for each model can be found in **Section S1** of the supporting information (SI). The receiver operating characteristic (ROC) curve, confusion matrix (CM), ROC-area under curve (AUC), Precision, Recall, and F1 scores are employed to evaluate the performance of the models. We also visually compare the model predictions with the ground truth of the processed μ-CT data for the test set. The Gini importance [21,22] is examined based on the RF model to access the feature importance to investigate how the neighboring voxels influence the state of a voxel. MLP is implemented in PyTorch [23], and all the other models are implemented in scikit-learn [24].

## 3. Result and discussion

In **Table 1**, model accuracies for all models as characterized by the ROC-AUC are presented for different percentages of hold-out test data sizes. The highest ROC-AUC for each kernel size (i.e., $K_i$) is highlighted in bold. In **Table 1**, the ground truth of the number of defective voxels in the test set and that predicted from the models are also shown. Among all the models tested, KNN and RF consistently out-perform other models regardless of the percentage of hold-out data for testing and kernel size. DT models are close followers as reflected by the ROC-AUC scores, while the MLP, LR and AdaBoost models show inferior accuracies. Comparing different test data percentages shows that by decreasing the number of test data, the ROC-AUC generally increase for all ML models, because more data are used for training. For almost all models, by increasing the kernel size, which includes more neighboring voxel thermal features as inputs for prediction, the ROC-AUC scores increase (**Table 1**). For KNN, RF and DT models, which have high ROC-AUC scores, when the percentage of the hold-out test data becomes larger (e.g., 30% and 40%), the increasing trend in ROC-AUC as kernel size increases is more obvious. For the hold-out test data size less than 20%, the ROC-AUC scores for these three models are seen weakly increasing or fluctuating around a high value. For the rest of the models, the increasing trend is evident regardless of the hold-out test data size. Overall, these finding suggest that neighboring voxels contains useful information that can help determine the state of the voxel of interest. This improvement in ROC-AUC is achieved despite the fact that increasing the kernel size decreases the available data for training and testing since more voxels are considered surface voxels and are excluded (see **Figure 1c**). Since RF turns out to be the best performing model, we will focus on analyzing the RF model in the rest of the paper.

**Table 1.** Model prediction and ROC-AUC for six ML models using 40 %, 30 %, 20 %, 10 % and 7 % of the data as the unseen test data and different kernel sizes

|  | Number of defective voxels in test set | | | | ROC-AUC | | | |
| --- | --- | --- | --- | --- | --- | --- | --- | --- |
|  | K1 | K3 | K5 | K7 | K1 | K3 | K5 | K7 |
| 7% hold-out data as test set | | | | | | | | |
| **Ground truth** | 962 | 746 | 485 | 285 | | | | |
| **KNN** | 1003 | 805 | 550 | 327 | **0.991** | **0.991** | **0.991** | 0.991 |
| **RF** | 1021 | 785 | 507 | 301 | 0.990 | **0.991** | **0.991** | **0.992** |

| Model | | | | | | | | |
|---|---|---|---|---|---|---|---|---|
| DT | 1013 | 801 | 516 | 308 | 0.986 | 0.985 | 0.984 | 0.985 |
| MLP | 3908 | 2171 | 1258 | 663 | 0.778 | 0.842 | 0.867 | 0.911 |
| LR | 3424 | 2124 | 1427 | 810 | 0.770 | 0.815 | 0.844 | 0.871 |
| AdaBoost | 3485 | 2088 | 1365 | 843 | 0.771 | 0.805 | 0.829 | 0.848 |
| **10% hold-out data as test set** | | | | | | | | |
| Ground truth | 1380 | 1081 | 690 | 420 | | | | |
| KNN | 1477 | 1195 | 820 | 508 | **0.988** | **0.987** | **0.987** | 0.987 |
| RF | 1520 | 1153 | 734 | 454 | 0.987 | **0.987** | **0.987** | **0.988** |
| DT | 1491 | 1175 | 751 | 458 | 0.983 | 0.979 | 0.977 | 0.981 |
| MLP | 5439 | 3008 | 1931 | 975 | 0.773 | 0.843 | 0.866 | 0.910 |
| LR | 4896 | 3045 | 2031 | 1166 | 0.760 | 0.816 | 0.841 | 0.855 |
| AdaBoost | 5210 | 2971 | 1954 | 1220 | 0.767 | 0.811 | 0.826 | 0.833 |
| **20% hold-out data as test set** | | | | | | | | |
| Ground truth | 2793 | 2101 | 1404 | 841 | | | | |
| KNN | 3244 | 2591 | 1923 | 1198 | **0.974** | **0.975** | 0.974 | 0.976 |
| RF | 3370 | 2435 | 1591 | 975 | 0.968 | 0.973 | **0.977** | **0.979** |
| DT | 3298 | 2471 | 1656 | 993 | 0.964 | 0.960 | 0.961 | 0.961 |
| MLP | 10868 | 6160 | 3938 | 1963 | 0.772 | 0.846 | 0.860 | 0.900 |
| LR | 9769 | 6149 | 4073 | 2331 | 0.761 | 0.815 | 0.840 | 0.852 |
| AdaBoost | 9831 | 6016 | 3949 | 2384 | 0.765 | 0.814 | 0.825 | 0.840 |
| **30% hold-out data as test set** | | | | | | | | |
| Ground truth | 4230 | 3158 | 2078 | 1265 | | | | |
| KNN | 5306 | 4365 | 3286 | 2056 | **0.954** | **0.963** | 0.963 | 0.964 |
| RF | 5619 | 3912 | 2497 | 1561 | 0.948 | 0.961 | **0.966** | **0.974** |
| DT | 5327 | 4058 | 2633 | 1644 | 0.938 | 0.944 | 0.942 | 0.945 |
| MLP | 16358 | 8688 | 5373 | 3017 | 0.771 | 0.852 | 0.864 | 0.894 |
| LR | 14734 | 9126 | 6021 | 3511 | 0.758 | 0.821 | 0.838 | 0.848 |
| AdaBoost | 15715 | 8974 | 5835 | 3455 | 0.766 | 0.811 | 0.830 | 0.836 |
| **40% hold-out data as test set** | | | | | | | | |
| Ground truth | 5644 | 4226 | 2775 | 1714 | | | | |
| KNN | 7739 | 6477 | 5001 | 3178 | **0.940** | **0.950** | 0.950 | 0.951 |
| RF | 8246 | 5665 | 3607 | 2272 | 0.928 | 0.950 | **0.960** | **0.961** |
| DT | 7848 | 5880 | 3805 | 2358 | 0.916 | 0.920 | 0.922 | 0.930 |
| MLP | 21921 | 10985 | 7439 | 2629 | 0.770 | 0.844 | 0.868 | 0.870 |
| LR | 19652 | 12158 | 8078 | 4689 | 0.757 | 0.822 | 0.835 | 0.847 |
| AdaBoost | 19476 | 11934 | 7691 | 4625 | 0.762 | 0.811 | 0.821 | 0.836 |

We should note that since the data is highly imbalanced, the ROC-AUC can be misleading. For example, since our labels consist of ~92% healthy and ~8% defective states, we can have a large number of false positive (FP) predictions but still obtain relatively high ROC-AUC score. As a result, we examine the confusion matrices. **Figure 2** shows the confusion matrices of the RF model high-performing models with different kernel sizes and different hold-out data sizes. The confusion matrices show that the RF model predicts more FP (upper right quad in the matrix) than false negative (FN, lower left quad) incidences. This means that models, when they are wrong in prediction, tend to predict normal voxels as defective one, but are less likely to miss defective voxels. This suggests that the models are "safe" but "conservative", which are favorable for practical AM applications by providing a larger margin of safety. The same observation can be made for the other two high-performance ML models (i.e., KNN and DT), see **Section S2** of the SI.

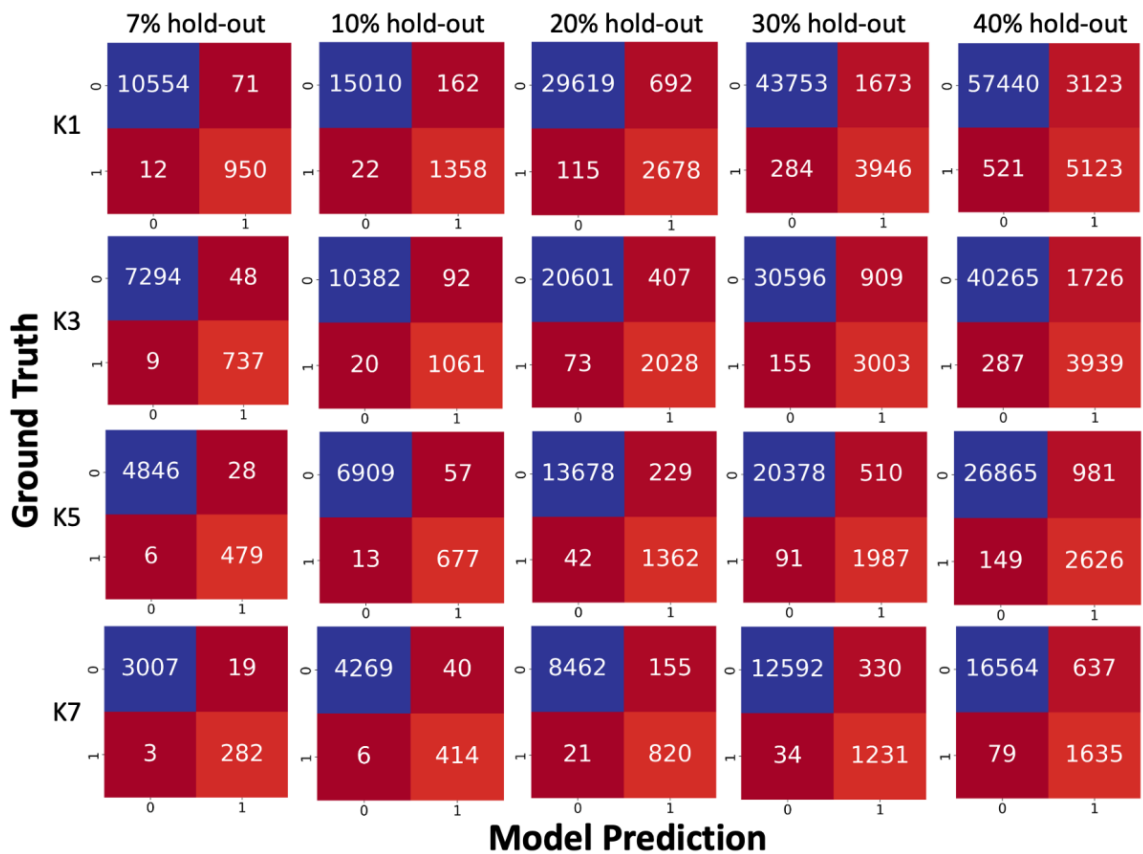

**Figure 2.** Confusion matrices for the RF models on different hold-out test data using different kernels.

**Table 2** shows the Precision (Pr), Recall (Re), F1 (i.e., 2×(Pr×Re)/(Pr+Re)) and accuracy (i.e., (TP+TN)/(TP+FN+FP+TN)) scores for the RF models. As shown in **Table 2**, by increasing the size of hold-out data, Pr, Re, F1 and accuracy scores all decrease. This is because less data is used for training when the hold-out test set size increases. By increasing the kernel size, the Pr score generally increases from K1 to K5, but it sees slight decrease in some cases when changing from K5 to K7. The Re score generally increases when the kernel size increases and this increase is more significant for the larger hold-out test set. One exception is the 10% hold-out case where the Re score fluctuates without a monotonic trend. The Re scores are uniformly higher than their Pr counterparts, which is due to the fact that the model predicts more FP than FN as discussed previously. The F1 score, which is a combined description of Pr and Re, almost always increase from K1 to K5, but it stagnates or decreases slightly when the kernel size increases to K7. A similar trend can be found in the accuracy score. From these quantifications, it can be inferred that there is a gain of useful information by including thermal information from more neighboring voxels up to K5 (i.e., the 3$^{rd}$ nearest neighbors), which increases the model quality. However, using the K7 kernel no longer improves model accuracy, suggesting that the thermal history of the 4$^{th}$ nearest neighboring voxels have little impact to the state of the present voxel.

**Table 2.** Precision Score, recall score, F1 score and accuracy score for the RF model on different hold-out size as the unseen test data and different kernel sizes.

| Hold-out test size | Precision Score (Pr) | | | | Recall Score (Re) | | | | F1-Score | | | | Accuracy Score | | | |
|---|---|---|---|---|---|---|---|---|---|---|---|---|---|---|---|---|
| | K1 | K3 | K5 | K7 | K1 | K3 | K5 | K7 | K1 | K3 | K5 | K7 | K1 | K3 | K5 | K7 |
| 7% | 0.930 | 0.939 | 0.945 | 0.937 | 0.988 | 0.988 | 0.988 | 0.990 | 0.958 | 0.963 | 0.966 | 0.962 | 0.993 | 0.993 | 0.994 | 0.993 |
| 10% | 0.893 | 0.920 | 0.922 | 0.912 | 0.984 | 0.981 | 0.981 | 0.986 | 0.937 | 0.950 | 0.951 | 0.947 | 0.989 | 0.990 | 0.991 | 0.990 |
| 20% | 0.795 | 0.833 | 0.856 | 0.841 | 0.959 | 0.965 | 0.970 | 0.975 | 0.869 | 0.894 | 0.910 | 0.903 | 0.976 | 0.979 | 0.982 | 0.981 |
| 30% | 0.702 | 0.768 | 0.796 | 0.789 | 0.933 | 0.951 | 0.956 | 0.973 | 0.801 | 0.850 | 0.869 | 0.871 | 0.961 | 0.969 | 0.974 | 0.974 |
| 40% | 0.621 | 0.695 | 0.728 | 0.720 | 0.908 | 0.932 | 0.946 | 0.954 | 0.738 | 0.796 | 0.823 | 0.820 | 0.945 | 0.956 | 0.963 | 0.962 |

In **Figure 3**, we further show selected Precision-Recall curves of models with hold-out data percentages of 10% and 30% to provide a visual comparison between the three high-performance models. The curve stretched towards the upper right corner in the Precision-Recall indicate better model prediction quality. As seen, the RF model is comparatively better than the KNN and DT models. It is important to see from **Figures 3c** and **3d** that when the K1 kernel is used (i.e., no neighboring voxel information used in prediction), the Precision-Recall curve is significantly inferior compared to the K5 kernel, again highlighting the importance of the thermal features from neighboring voxels in predicting the state of the central voxel. From the above discussion, we can again conclude that RF is the best performing model.

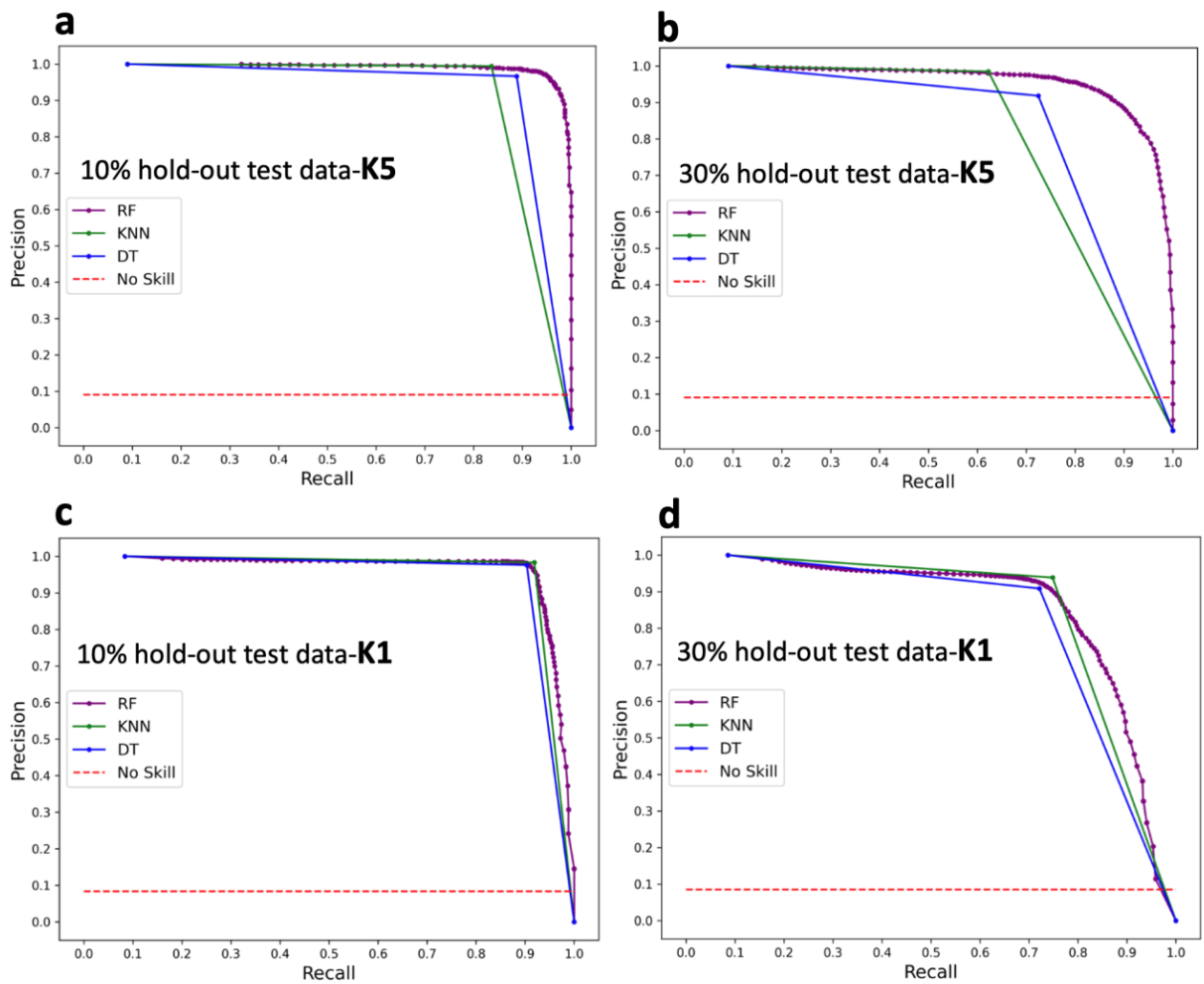

**Figure 3.** Precision-Recall curves for the RF, KNN and DT models using a K5 kernel for (**a**) 10% hold-out test data and (**b**) 30% hold-out test data, and those for the K1 kernel are shown in (**c**) and (**d**).

In **Figure 4**, we visually compare the predictions from the RF model and the ground truth for the 30% hold-out test data case in the 3D view and from randomly selected layers. The RF model shown uses the K5 kernel and the dots shown are the test data, *i.e.*, training data is not shown. As can be seen, both the 3D view (**Figures 4a** and **4b**) and the randomly selected layers (**Figure 4c**) shows good agreement between ground truth and RF prediction with a relatively small number of false predictions (circled). More visual comparisons are shown in **Section S3** in the SI.

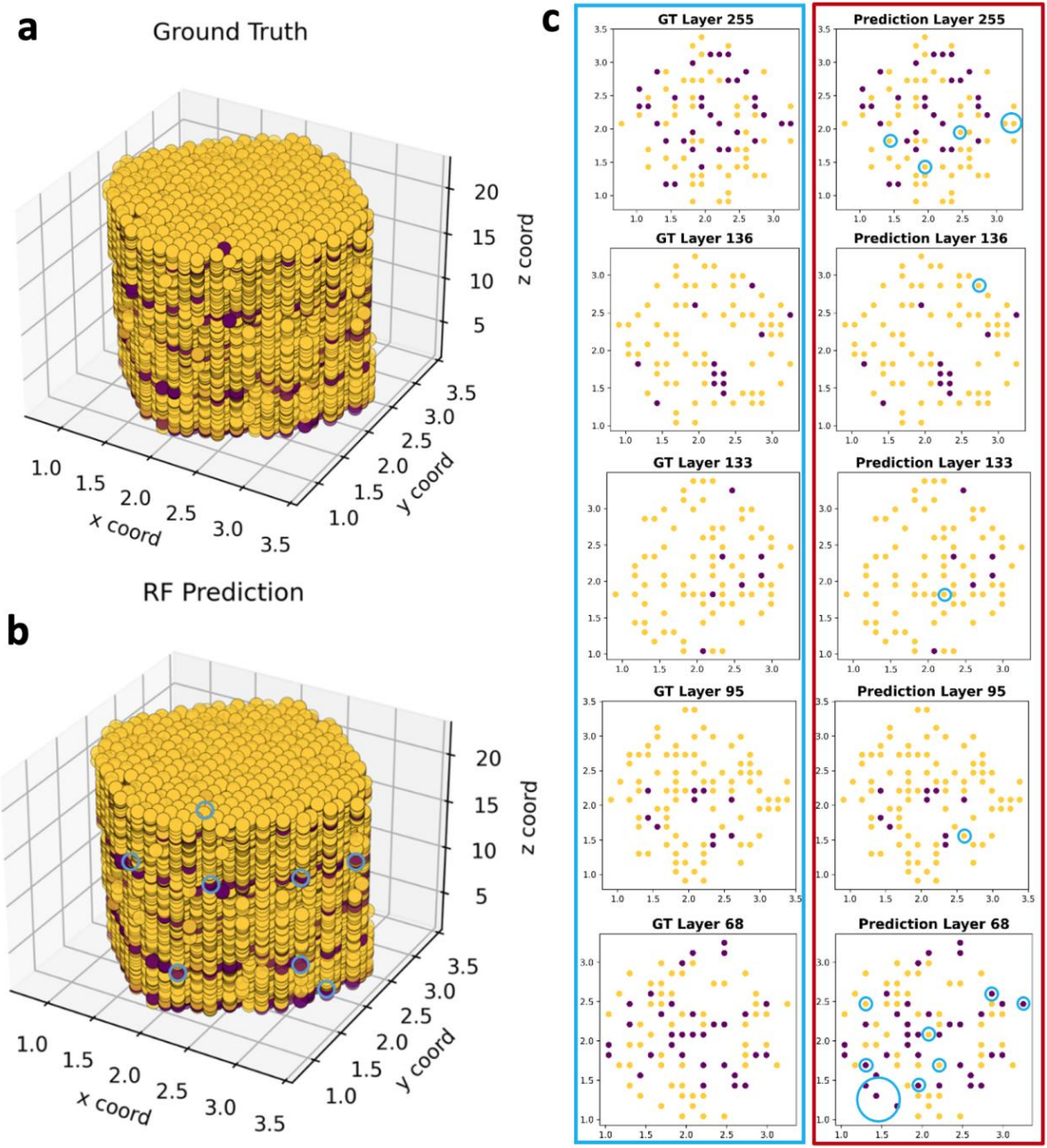

**Figure 4**. Visual comparison between the ground truth and RF prediction on the state of the test data set: the 3D view (coordinates in the unit of mm) of **(a)** the ground truth, and **(b)** the RF prediction, and **(c)** the side-by-side comparison of randomly selected layers. The blue circles indicate false predictions.

Lastly, we study the feature importance from the RF model with the K5 kernel to access the impact of the thermal feature of the neighboring voxels on the state of the central voxel. **Figure 5** shows the plot of the mean Gini feature importance score for the case with 30% hold-out test data. The training data is used to obtain the scores of the features based on the Gini impurity. For computing this, the feature score values corresponding to 100 random split of 70%-30% train-test data for all 250 features in the K5 kernel are calculated and then the average of these scores are taken and shown. We note that the first and second 125 feature scores are corresponding to the time-above threshold ($\tau$) and the peak radiance ($T_{max}$), respectively. The $T_{max}$ feature score values are much higher than the $\tau$ feature score values, suggesting that $T_{max}$ is a more important feature, which is consistent to a recent study [19]. We have further colored the features from voxels belonging to different layers in **Figure 5** as L+2, L+1, L0, L-1 and L-2, where L0 is the present layer, and "+" indicates layers above the present while "-" indicates those below it. The average feature scores of each layer are also shown in the figure. It is understandable that L0 has the highest average feature score of 0.032. L+2 has the second highest of 0.024, while L+1 has the third highest of 0.021. The features in L-1 and L-2, which are layers below the present layer, are not as important as those above it. These findings suggest that the RF model correctly captures the heat transfer nature involved in the LPBF process, where thermal energy in the top layer transferring downwards can impact the microporosity of layers beneath it. This is because when pores are formed due to lack of fusion in a region, its thermal conductivity is lower than normal regions, which will in turn raises the temperature of layers above it when the laser scans. For keyhole mode, the higher temperature in a layer generates pores below the current surface. Both scenarios suggest that one would see a much greater effect of the thermal process in layers above the present layer than those below.

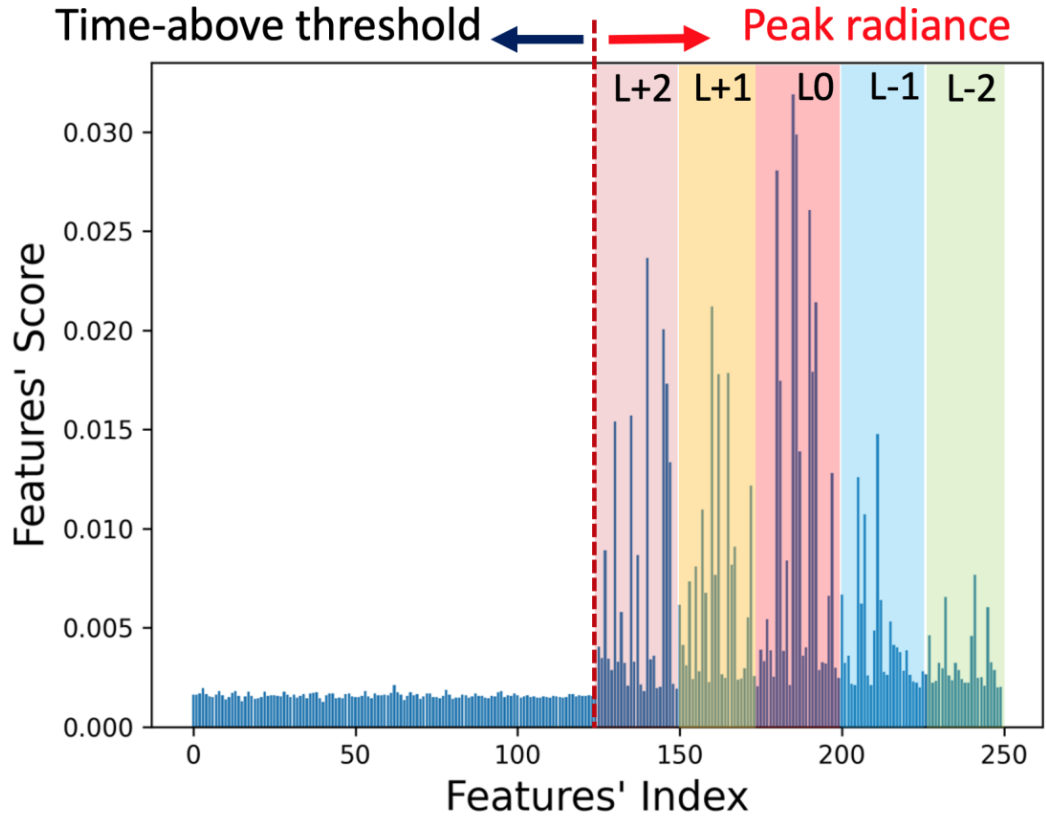

**Figure 5.** Mean of the features' score vs. features' index corresponding to the 5×5×5 voxel.

## 4. Conclusion

In this work, we develop machine learning models that can use *in-situ* thermographic data to predict the voxel state of LPBF stainless steel materials. Two features from the thermograph, including the time above the apparent melting threshold ($\tau$) and the maximum radiance ($T_{max}$) of each voxel, are used as inputs, and the binary state of each voxel, either defective or normal, is the output. Six different ML models are trained and tested for the binary classification task, with RF showing the highest predictive performance with the highest F1 score reaching 0.966. Besides using the thermal features of each voxel to predict its own state, those of neighboring voxels are also included, which is shown to improve the prediction accuracy. Such a finding suggests that the thermal history around a voxel can also impact its state due to heat transfer. Results also show that useful information can be collected from up to the 3$^{rd}$ nearest neighbor to improve model accuracy, but including scope beyond the 3$^{rd}$ nearest neighbor shows limited effect. Feature importance analysis based on the RF model shows that $T_{max}$ is more important to the voxel state than $\tau$. The analysis also finds that the thermal history of the voxels above the present voxel is more influential than those beneath it, which agrees with the heat transfer nature in the LPBF process. Our study demonstrates the viability to use *in-situ* thermographic data to predict porosity in LPBF materials. Since ML models are fast, they may play integral roles in the optimization and control of such AM technologies.

## Acknowledgements

The authors are grateful for the data provided by Honeywell Federal Manufacturing and Technologies and obtained during work funded by Contract No. DE-NA0002839 with the U.S. Department of Energy. The computations are supported by the University of Notre Dame, Center for Research Computing (CRC).